\documentclass{article} 

\usepackage{skyfall}

\newcommand{\bb}[1]{\left( #1 \right)}

\newcommand{\diag}[1]{\mathrm{diag}\bb{#1}}

\newcommand{\Nc}{\mathcal{N}}

\newcommand{\Sc}{\mathcal{S}}
\newcommand{\Tc}{\mathcal{T}}

\newcommand{\Nb}{\mathbb{N}}

\newcommand{\Rb}{\mathbb{R}}

\newcommand{\thetav     }{\boldsymbol \theta     }

\newcommand{\sigmav     }{\boldsymbol \sigma     }

\newcommand{\phiv       }{\boldsymbol \phi       }

\ifx\BlackBox\undefined
\newcommand{\BlackBox}{\rule{1.5ex}{1.5ex}}  
\fi
\ifx\QED\undefined
\def\QED{~\rule[-1pt]{5pt}{5pt}\par\medskip}
\fi
\ifx\proof\undefined
\newenvironment{proof}{\par\noindent{\em Proof:\ }}{\hfill\BlackBox\\}
\fi
\ifx\theorem\undefined
\newtheorem{theorem}{Theorem}
\fi
\ifx\example\undefined
\newtheorem{example}{Example}
\fi
\ifx\property\undefined
\newtheorem{property}{Property}
\fi
\ifx\lemma\undefined
\newtheorem{lemma}{Lemma}
\fi
\ifx\proposition\undefined
\newtheorem{proposition}{Proposition}
\fi
\ifx\fact\undefined
\newtheorem{fact}{Fact}
\fi
\ifx\remark\undefined
\newtheorem{remark}{Remark}
\fi
\ifx\corollary\undefined
\newtheorem{corollary}{Corollary}
\fi
\ifx\definition\undefined
\newtheorem{definition}{Definition}
\fi
\ifx\conjecture\undefined
\newtheorem{conjecture}{Conjecture}
\fi
\ifx\axiom\undefined
\newtheorem{axiom}[theorem]{Axiom}
\fi
\ifx\claim\undefined
\newtheorem{claim}[theorem]{Claim}
\fi
\ifx\assumption\undefined
\newtheorem{assumption}{Assumption}
\fi
\ifx\question\undefined
\newtheorem{question}{Question}
\fi
\ifx\problem\undefined
\newtheorem{problem}{Problem}
\fi

\date{}

\usepackage[T1]{fontenc}
\usepackage{lmodern}
\crefname{appx}{Appx}{Appx}
\Crefname{appx}{Appx}{Appx}
\crefname{section}{Sec}{Sec}
\Crefname{section}{Sec}{Sec}
\crefname{figure}{Fig}{Fig}
\Crefname{figure}{Fig}{Fig}

\DefineVerbatimEnvironment{UserTurn}{Verbatim}{
  fontsize=\small,
  frame=single,
  rulecolor=\color{OliveGreen},
  framerule=1pt,        
  framesep=3pt,         
  baselinestretch=1.0,
  xleftmargin=1em,
  xrightmargin=1em,
  breaklines=true,
  breakanywhere=true,
}

\DefineVerbatimEnvironment{ModelTurn}{Verbatim}{
  fontsize=\small,
  frame=single,
  rulecolor=\color{gray},
  framerule=1pt,        
  framesep=3pt,         
  baselinestretch=1.0,
  xleftmargin=1em,
  xrightmargin=1em,
  breaklines=true,
  breakanywhere=true,
}

\title{SCOPE: Language Models as One-Time Teacher for \\ Hierarchical Planning in Text Environments}

\author{
Haoye Lu, Pavan Seshadri, Kaheer Suleman\\
\texttt{\{haoye,pavan.seshadri,kaheer\}@skyfall.ai}
}

\newcommand{\EWM}{\text{EWM}}
\newcommand{\MWM}{\text{MWM}}

\begin{document}

\begin{textblock*}{5cm}(1.5cm,2.2cm)
    \includegraphics[width=4.2cm]{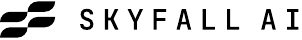}
\end{textblock*}

\maketitle

\begin{abstract}
Long-term planning in complex, text-based environments presents significant challenges due to open-ended action spaces, ambiguous observations, and sparse feedback. Recent research suggests that large language models (LLMs) encode rich semantic knowledge about the world, which can be valuable for guiding agents in high-level reasoning and planning across both embodied and purely textual settings. However, existing approaches often depend heavily on querying LLMs during training and inference, making them computationally expensive and difficult to deploy efficiently. In addition, these methods typically employ a pretrained, unaltered LLM whose parameters remain fixed throughout training, providing no opportunity for adaptation to the target task. To address these limitations, we introduce SCOPE (Subgoal-COnditioned Pretraining for Efficient planning), a one-shot hierarchical planner that leverages LLM-generated subgoals only at initialization to pretrain a lightweight student model. Unlike prior approaches that distill LLM knowledge by repeatedly prompting the model to adaptively generate subgoals during training, our method derives subgoals directly from example trajectories. This design removes the need for repeated LLM queries, significantly improving efficiency, though at the cost of reduced explainability and potentially suboptimal subgoals. Despite their suboptimality, our results on the TextCraft environment show that LLM-generated subgoals can still serve as a strong starting point for hierarchical goal decomposition in text-based planning tasks. Compared to the LLM-based hierarchical agent ADaPT \citep{PrasadKHCSBK2024}, which achieves a 0.52 success rate, our method reaches 0.56 and reduces inference time from 164.4 seconds to just 3.0 seconds.
\end{abstract}

\section{Introduction}
Developing agents capable of planning and reasoning over long horizons remains a central challenge in reinforcement learning (RL). Classical RL and planning methods have achieved notable success in environments with compact state representations and finite action spaces \citep{BercherAL2019, PateriaSTQ2021}. However, such favourable conditions typically rely on substantial domain expertise to design effective state abstractions, carefully shaped rewards, and constrained action sets \citep{IbrahimMJSO2024}. Extending these successes to more general and unstructured settings is considerably more difficult, as many real-world problems lack such engineered structure, leading to persistent challenges in exploration, credit assignment, and generalization.

Motivated by the empirical success of large language models (LLMs) and their strong capability for general-purpose text processing, recent RL research has begun leveraging their semantic knowledge to support high-level reasoning and planning in embodied agents. This approach has shown strong promise in both robotic control \citep{AhnBBCC2022} and long-term planning in open-world environments such as Minecraft \citep{WangCMLJLM2024}. However, because these methods rely heavily on querying LLMs during both training and inference, they incur high computational costs and are difficult to deploy efficiently. Furthermore, although some approaches introduce additional trainable components to give the LLM-based planner limited adaptability, the LLM itself is typically kept frozen, preventing its parameters from being updated to better fit the target task. 

To address the high computational cost and limited flexibility of LLM-based planners, \citet{ZhouHZZL2024, LiPSFXZ2026} propose an alternative approach in which the LLM serves as a teacher to supervise a small student network for high-level decision making. Early in training, the student closely follows the LLM’s guidance, but its reliance gradually diminishes, eventually allowing it to operate independently at inference time. Although this approach removes the need for the LLM during deployment, it still requires iterative LLM queries throughout training to provide ongoing guidance to the student.

Adopting a similar idea, in this work, we propose SCOPE (Subgoal-COnditioned Pretraining for Efficient planning), a more efficient approach that uses LLM-generated subgoals only once at initialization, assuming access to a set of suboptimal example trajectories. Unlike prior methods that repeatedly query the LLM to generate subgoals during training adaptively, SCOPE directly extracts subgoal sequences from demonstrations and uses them to pretrain a student planner. Although these subgoals are suboptimal as the LLM does not interact with the environment, our preliminary results on TextCraft (a simplified, text-only version of Minecraft) show that they still provide a strong starting point for the planner in text-based planning tasks. After fine-tuning the planner by maximizing expected return through interaction with a world model, SCOPE achieves strong performance. In particular, compared to the LLM-based hierarchical agent ADaPT \citep{PrasadKHCSBK2024}, which obtains a 0.52 success rate on TextCraft, SCOPE reaches 0.56. It is also significantly more efficient at inference time: SCOPE completes the game in an average of 3.0 seconds on a single NVIDIA A10 GPU, whereas ADaPT requires 164.4 seconds using a GPT-3.5 backend accessed via the OpenAI API under ideal network conditions.

\section{Related Work}

The hierarchical paradigm for planning and executing long-term tasks has been extensively explored over the past several decades \citep{BercherAL2019, PateriaSTQ2021}. The central idea is to decompose complex, long-horizon decision-making problems into a hierarchy of simpler subtasks, where a high-level policy operates over abstract actions, selecting subtasks rather than primitive controls, to achieve the overall objective. Each subtask can itself be formulated as a reinforcement learning problem, solved by a lower-level policy that learns to accomplish it \citep{Hengst2010}. The interaction between these hierarchical policies collectively governs the agent's overall behaviour.

By representing long-horizon problems in terms of temporally extended subtasks, hierarchical reinforcement learning and planning effectively shortens the decision horizon, a principle known as temporal abstraction \citep{BartoM2003, Dietterich2000, SuttonPS2000}. Temporal abstraction enables more efficient credit assignment across extended timescales \citep{VezhnevetsOSHJSK2017}, while decomposing tasks into subtasks typically simplifies learning and promotes more structured exploration during training \citep{NachumTLGLL2019}. These advantages make HRL a powerful framework for scaling reinforcement learning to long-horizon domains \citep{DayanH1992, VezhnevetsOSHJSK2017, NachumGLN2018}. Empirical evidence shows that HRL consistently outperforms flat RL methods across diverse settings, including continuous control \citep{FlorensaDA2017, LevyPS2018}, strategic games \citep{VezhnevetsOSHJSK2017, NachumGLN2018}, and robotic manipulation \citep{FoxKSG2017, GuptaKLLH2019}. Notably, several studies attribute these improvements primarily to enhanced exploration facilitated by subgoal-based hierarchical structures \citep{JongHST2008, NachumTLGLL2019}.

Given that large language models (LLMs) encode rich semantic knowledge about the world, such knowledge can be highly valuable for guiding embodied agents to perform high-level reasoning and planning. Motivated by this insight, \citet{AhnBBCC2022} introduced a hierarchical planning framework for robot control, in which a low-level agent executes concrete motor actions, such as opening and closing the gripper, while a high-level, LLM-based agent issues abstract commands like ``put the Coke on the counter.'' Building on this idea, subsequent studies have extensively explored the use of LLMs as task planners. These approaches can be broadly categorized into two types: comprehensive \citep{TangYTZFH2023, DalalCCS2024}, where all subgoals are planned in advance, and incremental, where subgoals are generated adaptively, allowing the planner to make dynamic adjustments based on feedback or environmental changes \citep{IchterBCKFHP2023, ZhangHL2023, WangCMLJLM2024}.


Due to the vast number of trainable parameters in LLM-based agents, directly optimizing them through fine-tuning is often impractical and computationally demanding \citep{HuHLKITYXYL2024,SchoeppJCYAGSZT2025}. Consequently, most existing approaches improve the behaviour of LLM agents indirectly, by refining their prompts based on environmental feedback and/or by allowing the LLM to generate multiple high-level action candidates, from which a separate value function selects the most appropriate one \citep{AhnBBCC2022, WangCaiCLJLM2023}. Beyond this common paradigm, \citet{ZhouHZZL2024,LiPSFXZ2026} propose an alternative approach that leverages the LLM as a teacher to guide a smaller student network in producing high-level decisions. Initially, the agent is trained to follow the LLM's guidance closely; however, as training proceeds, its reliance on the LLM gradually diminishes, enabling it to make independent decisions. This design removes the need for the LLM during inference, allowing the student network to be further fine-tuned in the final training stage by directly maximizing the expected return.



\section{Backgrounds}
\label{sec:background}



\begin{figure*}[!t]
    \centering
    \includegraphics[width=0.8\textwidth]{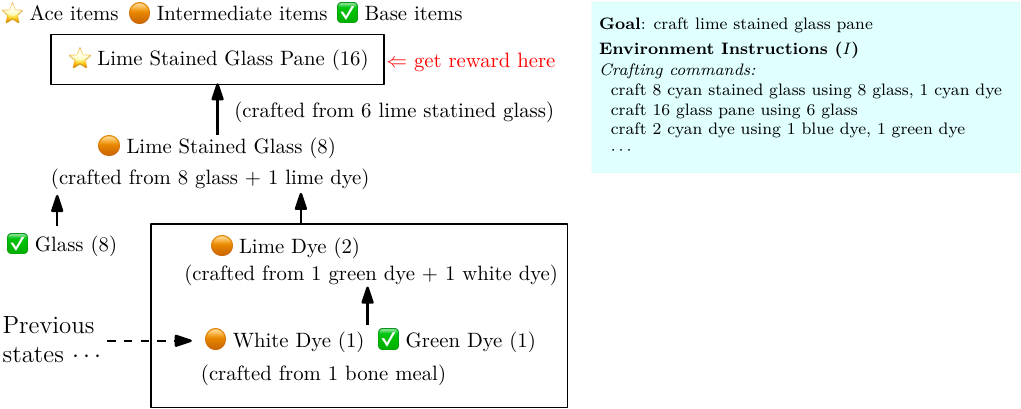}
    \caption{TextCraft Environment. An example crafting dependency chain for producing the ace item (lime stained glass pane). The agent must gather base items, synthesize intermediate items through the provided crafting commands, and execute the sequence in the correct order to obtain the final reward.
}
    \label{fig:textcraft}
\end{figure*}

In this work, we aim to develop an agent capable of solving long-horizon, goal-reaching tasks in a text-based environment, with only one-time guidance from an LLM prior to any agent training. Specifically, we consider a setting in which the system receives a user-provided instruction that includes a goal  \( g \) that the agent should accomplish, along with additional descriptions of the environment configuration \( I \). The agent receives a reward only upon successful completion of the task. We also assume access to a set of successful example goal-instruction-trajectory tuples,
\(\Sc = \{(g_k, I_k, T_k) \mid k = 1,2,\ldots,N\}\), provided by humans to help the agent learn how to interact with the environment. Each trajectory is defined as
\(T_k = (s_0, a_0, s_1, \ldots, a_{M_k-1}, s_{M_k})\), where each \(a_i\) is the action executed when the agent is in state \(s_i\), thereby forming a sequence of state–action transitions.
Although all example trajectories are assumed to eventually reach the final goal, we do not assume they are optimal, as human demonstrations may include suboptimal or inefficient actions.


\textbf{Language Models as One-Time Teacher.} To establish an initial foundation for goal decomposition, an LLM-based planner analyzes the demonstration trajectories and heuristically partitions them into subtrajectories, each ending at a state where a subgoal is deemed achieved. We then train a low-level employee agent to accomplish these short-horizon subgoals, while a high-level manager agent is responsible for proposing them.

To ensure consistent goal decomposition across all trajectories, rather than directly asking the LLM planner to output subtrajectories, we provide it with 50 randomly sampled instruction-trajectory pairs and ask it to propose a Python function \( f_{\text{dc}}(T) \) that systematically decomposes a trajectory \( T  = (s_0, a_0, s_1, \ldots, a_{M-1}, s_M)\) into subtrajectories with associated subgoals. This function is paired with a corresponding subgoal completion function \( f_{\text{sg}}(s, \tilde g) \), which returns \( 1 \) if the subgoal \( \tilde g \) is achieved in state \( s \), and \( 0 \) otherwise. Although the resulting subgoals may be suboptimal and lack interpretability, our empirical study demonstrates that they nevertheless serve as a strong initialization for subgoal decomposition and provide effective guidance for improving the efficiency of long-term planning training. (The prompts used and the generated Python programs are included in \cref{appx:llm_prompt_outputs}.) In pariticular, by applying \( f_{\text{dc}} \), we decompose tuple $(g, I, T)$ into 
\begin{align}
	(\tilde T_k, \tilde g_k, I) \text{\quad with \quad} \tilde T_k = (s_{i_{k-1}}, a_{i_{k-1}}, \ldots, s_{i_{k}}), \quad k = 1, 2, \ldots, N_T, \label{eq:def_subtraj}
\end{align}
where \( i_k \) denotes the index of the state at which subgoal \( \tilde{g}_k \) is achieved (i.e., \( f_{\text{sg}}(s_{i_k}, \tilde{g}_k) = 1 \)), with \( i_0 = 0 \). Here, \(N_T\) denotes the total number of subgoals, and completing the final subgoal \(\tilde{g}_{N_T}\) corresponds to successfully achieving the ultimate goal \(g\).  

For later reference, each subtrajectory in \cref{eq:def_subtraj} is further decomposed into state-action-subgoal-instruction tuples of the form
\((s_{i_{k-1}}, a_{i_{k-1}}, \tilde{g}_k, I)\), which are compiled into the dataset
\(\Phi = \{(s, a, \tilde{g}, I)\}\). In addition, for tuple $(g, I, T) \in \Sc$, let 
\begin{align}
    \Phi_0^{g, I, T} = \big\{\, (s_{i_{k-1}}, \tilde{g}_k, g, I) \;:\; k = 1, 2, \ldots, N_T \,\big\},
\end{align}
where each element of $\Phi_0^{g, I, T}$ corresponds to the initial state of a subtrajectory induced from $T$, paired with its associated subgoal $\tilde{g}_k$, ultimate goal $g$ and instruction $I$. We further define
\begin{align}
	\Phi_0 = \bigcup_{(g, I,T) \, \in \, \Sc} \Phi_0^{g, I, T}. 
\end{align}
\textbf{TextCraft Environment.} In this preliminary work, we use the TextCraft environment \citep{PrasadKHCSBK2024} as the benchmark to evaluate our hierarchical text-based RL framework. TextCraft, inspired by Minecraft, tests an agent's ability to perform compositional reasoning and long-term planning in crafting tasks. As shown in \cref{fig:textcraft}, at the start of each episode, the agent receives a goal $g$ to craft a specific ice item as well as additional instructions $I$ containing a set of available crafting commands. The agent is expected to infer which base items are non-craftable and obtainable from the environment, then act through textual commands -- \texttt{craft} using \texttt{<ingredients>} to produce intermediate or final items, and \texttt{get} to collect the required base materials. The agent tracks its inventory through the textual description of the current state. An episode terminates once the target item is successfully crafted, at which point the agent receives a reward; otherwise, no reward is provided. 
Although the environment appears simple, it is intentionally designed so that crafting the final item requires executing a precise sequence of actions in the correct order, including the creation of multiple intermediate items. This makes TextCraft a suitable testbed for evaluating the effectiveness of hierarchical planning algorithms in an open, purely text-based setting.


\section{SCOPE}
\label{sec:method}
\textbf{Overview.} Similar to prior long-term planning work in open-world environments \citep{AhnBBCC2022, WangCMLJLM2024}, our approach, SCOPE, adopts a hierarchical agent design consisting of an employee agent for low-level execution and a manager agent for high-level planning. The manager agent proposes a high-level plan based on the current environment state and the ultimate goal, and then delegates control to the employee agent. The employee agent interacts directly with the environment to complete the assigned subgoals and returns control to the manager either upon completion or once a preset step limit is reached. This iterative process continues as the manager proposes subsequent subgoals, and terminates once the ultimate goal has been achieved. 

The remainder of this section describes how both agents are trained. Each agent is first pretrained on the suboptimal trajectories, followed by a reinforcement learning-based stage to further improve the goal-achievement rate. The RL training relies on interactions with world models. For the employee agent, the elementary world model is obtained by training directly on the provided suboptimal trajectories. The world model used for training the manager agent is then constructed by combining the trained employee agent with the elementary world model; its implementation details will be presented in the manager training section.

\subsection{Employee Agent}
The employee agent is responsible for accomplishing individual subgoals and is trained in two stages: an imitation-based pretraining phase to mimic suboptimal trajectories, followed by RL fine-tuning to maximize subgoal completion rate.

\textbf{Pretraining.} For pretraining, the agent $\pi^e_{\thetav}$  is trained to imitate suboptimal trajectories collected from human players by minimizing 
\begin{align}
	J^e(\thetav) = - \sum_{(s, a, \tilde g, I) \in \Phi} \log \pi^e_{\thetav}(a \vert s, \tilde g, I). 
\end{align}
This objective encourages the policy \(\pi^e_{\thetav}\) to reproduce the demonstrated behaviour, allowing it to learn an initial mapping from states and subgoals to actions under the given instruction.

\begin{figure*}
    \centering
    \includegraphics[width=0.9\textwidth]{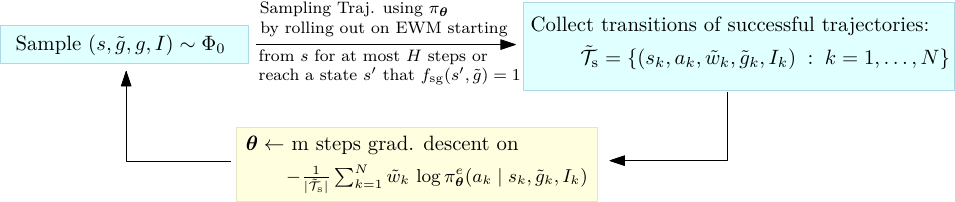}
    \caption{The RL training pipeline of the employee agent.}
    \label{fig:method}
\end{figure*}%
\textbf{Employee World Model.} The RL training builds on a world model trained from suboptimal trajectories to predict the next state and check action validity. Given the current state \(s\), action \(a\), and instruction \(I\), the elementary world model \(\EWM(s, a; I)\) first verifies whether \(a\) is valid. In TextCraft, an action is invalid if it cannot be parsed (e.g., semantically incorrect) or executed, for example, when a crafting command is missing from \(I\) or when the inventory lacks the required materials. If valid, the model predicts the next state; otherwise, it returns the current state, treating the action as rejected. It outputs both the next state and the validity flag. Since the state in TextCraft is represented as an inventory dictionary, we employ a neural network that takes \((s, a; I)\) in text form to predict only the action validity, while the next-state update is implemented directly through set operations.

\textbf{RL Finetuning.} In this stage, the employee agent is further trained to maximize the expected subgoal achievement rate by interacting with the world model  \(\EWM(s, a; I)\). We use $f_{sg}$ function provided by the LLM-planner to check if a subgoal is completed. In our implementation, we adopt the Cross-Entropy Method (CEM,  \citealt{MannorRG2003,deBoerKMR2005}), a widely used derivative-free model-based RL (MBRL) algorithm that has demonstrated strong performance across challenging benchmarks \citep{WangB2020,YangCIZTS2020}. As illustrated in \cref{fig:method}, at each iteration the employee agent samples a random tuple $(s, \tilde{g}, g, I) \in \Phi_0$ and performs a rollout in the world model starting from $s$, which results in a new trajectory by executing a sequence of actions sampled from its policy network, conditioned on the current state, subgoal $\tilde{g}$ and instruction $I$. A total of $\Nc_e$ trajectories are sampled per iteration.

If a trajectory reaches a state \(s'\) such that \(f_{\mathrm{sg}}(s',\tilde g)=1\) within a preset number of steps, then each state-action pair \((s,a)\) in that trajectory is assigned the normalized weight
\(
\tilde w = \frac{1}{\lvert \text{traj} \rvert},
\)
where \(\lvert \text{traj} \rvert\) denotes the number of transitions; otherwise, the weight is set to zero. This scheme penalizes unnecessarily long trajectories and encourages achieving the subgoal with fewer steps. Collecting all weighted state-action pairs from successful trajectories yields
\[
\tilde \Tc_{\mathrm{s}}
= \big\{ (s_k, a_k, \tilde w_k, \tilde g_k, I_k) \;:\; k=1,2, \ldots,  \big\}
\]
where each tuple contains a state-action pair together with its assigned weight \(\tilde w_k\), the associated subgoal \(\tilde g_k\), and instruction \(I_k\). If $\tilde \Tc_\mathrm{s}$ is non-empty, the policy parameters $\thetav$ are then updated by minimizing the weighted negative log-likelihood over $\tilde  \Tc_{\mathrm{s}}$:
\begin{equation}
-  \frac{1}{|\tilde  \Tc_{\mathrm{s}}|}\sum_k
\tilde w_{k} \, \log \pi^e_{\thetav}(a_k \mid s_k, \tilde g_k, I_k),
\end{equation}
for a preset number of gradient descent steps, where $|\tilde  \Tc_{\mathrm{s}}|$ denotes the total number of tuples in $\tilde \Tc_{\mathrm{s}}$; otherwise, the update is skipped.  The process is repeated by resampling a new set of $\Nc_e$ trajectories.

\subsection{Manager Agent}
Unlike the employee agent, which selects actions that directly interact with the environment, the manager agent proposes the next subgoal based on the current state and the ultimate goal, guiding the employee toward achieving that final goal.

\textbf{Pretraining.} For pretraining, the manager agent is trained to autoregressively reproduce the subgoal sequences generated by $f_{\text{dc}}(I, T)$ for $(g, I, T) \in \mathcal{S}$. This can be achieved by minimizing
\begin{align}
    \mathcal{J}^m(\boldsymbol{\phi})
    = - \sum_{(g, I, T) \in \mathcal{S}} \; \sum_{k = 1}^{N_T}
        ~~\log \pi_{\boldsymbol{\phi}}^{m} \big( \tilde{g}_k \mid s_{i_{k-1}}, g, I \big).
\end{align}
In this way, the manager agent acquires an initial ability to propose subgoals for each subtrajectory, conditioned on its initial state \(s_{i_{k-1}}\), the ultimate goal \(g\), and the accompanying instruction \(I\). This pretrained policy serves as a reasonable starting point, but is not expected to be optimal. 



After pretraining, the manager agent is further optimized using RL to maximize the ultimate goal-achievement rate. This stage involves rolling out trajectories within a world model that operates at the subgoal level. We show that this manager-level world model can be constructed by combining the employee world model \(\EWM\) with the trained employee agent \(\pi^{e}_{\thetav}\), which the manager needs to coordinate with during planning.

\textbf{Manager World Model.} In contrast to the employee world model, the manager world model treats subgoals as actions. Therefore, correspondingly, the manager world model $\MWM(s, \tilde{g})$ takes the current state $s$ and a proposed subgoal $\tilde{g}$ as input and first checks whether $\tilde{g}$ is achievable within a preset number of steps. If so, it returns the resulting state upon subgoal completion; otherwise, it returns $s$. Here, ``achievable'' means that the subgoal is both plausible and executable by the deployed employee agent. To implement this, we roll out the employee policy \(\pi_{\thetav}^e\) on the employee world model \(\EWM\) starting from \(s\). If \(\tilde{g}\) is reached within the step limit, the final state is returned; otherwise, the initial state is returned.

To determine whether the ultimate goal is achieved, we can use an ultimate-goal completion function $f_{\text{ug}}(s, g)$, which returns $1$ if the ultimate goal $g$ is satisfied in state $s$ and $0$ otherwise. In general, $f_{\text{ug}}$ can be implemented using a neural network trained via supervised learning on the provided suboptimal trajectories. In TextCraft, however, it can be computed directly by checking whether the inventory contains the ace item.

\textbf{RL Finetuning.} 
In the RL stage, the manager agent is trained in a manner similar to the employee agent, but operates at the subgoal-proposing level by interacting with $\MWM$. In particular, at each iteration, the manager agent samples a random tuple $(s,\tilde g, g, I) \in \Phi_0$ and performs a rollout in $\MWM$ starting from $s$.\footnote{During RL fine-tuning, subgoals are drawn from $\pi^m_{\phiv}(\cdot \mid s, g, I)$, and the $\tilde g$ in $\Phi_0$ is not used.} The rollout ends when either a preset maximum number of steps is reached or the ultimate goal $g$ is achieved. This produces a new state-subgoal trajectory $(s_0, \tilde g_0, s_1, \tilde g_1, \ldots, s_{l-1}, \tilde g_{l-1}, s_l)$, where $s_0 = s$ and $l$ is the total number of proposed subgoals. If $f_{\mathrm{ug}}(s_l, g) = 1$, the trajectory is considered successful and each state-subgoal pair $(s_k, a_k)$ for $k = 0,1,\ldots,l-1$ is assigned a normalized weight $w = \frac{1}{l}$. Otherwise, the trajectory is treated as a failure, and all state-subgoal pairs receive weight zero.  Similar to the RL finetuning of the employee agent, this scheme penalizes unnecessarily long trajectories and encourages achieving the ultimate goal with a smaller number of subgoals. At each iteration, $\Nc_m$ trajectories are generated. Gathering the weighted state-subgoal pairs from all successful trajectories yields
\[
\Tc_{\mathrm{s}}
= \big\{ (s_k, \tilde{g}_k, w_k, g_k, I_k) \;:\; k=1,2, \ldots  \big\},
\]
where each tuple contains a state-subgoal pair together with its assigned weight \(w_k\), the associated ultimate goal \(g_k\), and instruction \(I_k\). If $\Tc_\mathrm{s}$ is non-empty, the policy parameters $\phiv$ are then updated by minimizing the weighted negative log-likelihood over $\Tc_{\mathrm{s}}$:
\begin{equation}
-  \frac{1}{|\Tc_{\mathrm{s}}|}\sum_k
w_{k} \, \log \pi^m_{\phiv}(\tilde g_k \mid s_k, g_k, I_k),
\end{equation}
for a specified number of gradient updates, with $|\Tc_{\mathrm{s}}|$ indicating the number of tuples in the set. If $\Tc_{\mathrm{s}}$ is empty, no update is performed. The procedure then continues by sampling a fresh batch of $\Nc_m$ trajectories for the next iteration.

Upon completion of training, the manager and employee agents are combined in a hierarchical manner, as described earlier in this section, and deployed for evaluation in the environment. The effectiveness of this approach is demonstrated in \cref{sec:emp}.

\section{Empirical Results}
\label{sec:emp}
In this section, we conduct a preliminary empirical study of SCOPE using the TextCraft environment.

\textbf{Dataset and Network Architecture.} To simulate suboptimal trajectories resembling those of human players, we generate 500K rollouts, reserving 1K each for validation and testing. The objective is to obtain diverse trajectories that generally progress toward the target goal in the TextCraft environment while including occasional suboptimal actions to mimic human-like exploration. Starting from a given state space (defined by available crafting commands) and a target goal item, we construct the recipe tree, derive the optimal crafting sequence, and inject random actions at a 10\% rate to introduce variability. The resulting trajectories blend optimal and noisy transitions, producing behaviour that closely resembles human demonstrations.  

Although the employee and manager agents differ in their input-output formats, all signals are converted to text for unified processing, enabling both to share a variational sequence-to-sequence architecture (see \Cref{appx:model_arch} for details). The hyperparameter settings are provided in \Cref{appx:hyperparameter}. 

\begin{table*}
\centering
\caption{Success rates for crafting ace items in TextCraft. \textbf{Top:} ADaPT \citep{PrasadKHCSBK2024} is a hierarchical agent that uses a GPT-3.5-based planner \citep{BrownMRSSKD2020}. \textbf{Bottom:} Ablation results. SCOPE (hand-engineered-subgoal) replaces LLM-generated subgoals with manually constructed, interpretable ones. SCOPE (without manager RL-finetuning) removes the manager-level RL stage.
}
\label{tab:performance}
\begin{tabular}{l c c}
\toprule
\textbf{Setting} & \textbf{Success Rate} & \textbf{\# Parameters} \\
\midrule
ADaPT \citep{PrasadKHCSBK2024} & 0.52 & 175B \\
SCOPE (\textbf{ours})  & 0.56 & 11.04M \\
\midrule
SCOPE (hand-engineered-subgoal) & 0.58 & 11.04M \\
SCOPE (without manager RL-finetuning) & 0.24 & 11.04M \\
\bottomrule
\end{tabular}
\vspace{1em}  
\end{table*}

\begin{table*}[t]
    \centering
    \caption{Success rates for crafting ace items in TextCraft for ADaPT \citep{PrasadKHCSBK2024} with different backends. The original implementation uses GPT-3.5 \citep{BrownMRSSKD2020} as the backend.}
    \label{tab:backend-success}

\resizebox{\textwidth}{!}{
    \begin{tabular}{lccc}
    \toprule
    \textbf{Backend} & \textbf{Success Rate} & \textbf{\# Parameters} & \textbf{Open Weight?} \\
    \midrule
    GPT-4o \citep{OpenAI2024}                         & 0.58 & 1.8T$^{*}$ & No  \\
    Mistral Small 3 \citep{mistral_small3_2025}       & 0.58 & 24B        & Yes \\
    SCOPE (\textbf{ours})                          & 0.56 & 11.04M     & - \\
    GPT-3.5 \citep{BrownMRSSKD2020}                        & 0.52 & 175B       & No  \\
    GPT-4o mini \citep{OpenAI2024}                    & 0.43 & 8B$^{*}$   & No  \\
    DeepSeek-R1-Distill-Qwen-32B \citep{DeepSeek-AI2025}    & 0.13 & 32B        & Yes \\
    Claude-3 Haiku \citep{anthropic_claude3_2024}                 & 0.00 & 20B$^{*}$  & No  \\
    \bottomrule
\end{tabular} 
}%
    \vspace{0.3em}
    {\footnotesize $^{*}$ Rough estimates based on publicly discussed or widely believed parameter counts.}
\end{table*}


\subsection{Performance Comparison}
As shown in \Cref{tab:performance}, ADaPT \citep{PrasadKHCSBK2024} is a hierarchical framework that relies on an LLM-based planner with a GPT-3.5 backend \citep{BrownMRSSKD2020}, and it achieves a success rate of 0.52. In comparison, SCOPE attains an ultimate-goal success rate of 0.56. This shows that a one-time LLM initialization, followed by RL fine-tuning, can outperform methods that depend on LLM inference throughout execution.
The reduced model size also leads to substantially lower inference cost and latency. In particular, SCOPE agent completes the game in an average of 3.0 seconds on an NVIDIA A10 GPU, whereas ADaPT requires 164.4 seconds when using a GPT-3.5 backend accessed via the OpenAI API under ideal network conditions.

In addition to the original ADaPT implementation based on GPT-3.5 \citep{BrownMRSSKD2020}, we also report results using a broader selection of backends in \Cref{tab:backend-success}. The table compares ADaPT with a range of LLM backends that vary in size and openness. GPT-4o is a large proprietary model developed by OpenAI~(\citeyear{OpenAI2024}), and GPT-4o mini is a smaller, more cost-efficient variant intended to reduce latency and compute while retaining much of GPT-4o’s capability. Claude-3 Haiku \citep{anthropic_claude3_2024} is Anthropic’s lightweight model, designed to answer many requests quickly and reliably under tight latency or budget constraints. On the open-weight side, Mistral Small 3 \citep{mistral_small3_2025} is a 24B-parameter model in the ``small'' LLM category (below 70B) that delivers performance comparable to substantially larger models. DeepSeek-R1-Distill-Qwen-32B \citep{DeepSeek-AI2025} is a mid-sized reasoning model distilled from DeepSeek-R1 based on Qwen2.5 \citep{Qwen2024}. It achieves competitive or superior results to OpenAI o1-mini \citep{OpenAI2024-o1} on a range of benchmarks, representing a strong open-weight option in the dense-model regime. While using these more advanced backends improves ADaPT’s performance, SCOPE remains highly competitive: with only 11.04M parameters, it attains a success rate of 0.56, coming very close to the performance of GPT-4o and Mistral Small 3 (both 0.58 with tens of billions to trillions of parameters) and outperforming GPT-4o mini, DeepSeek-R1-Distill-Qwen-32B, and Claude-3 Haiku. This underscores how parameter-efficient SCOPE is compared to both closed and open-weight alternatives.

\subsection{Impact of Suboptimal Subgoals on Performance} 
\label{sec:impact_subopt_subgoal_performance}

To better understand how much performance degradation results from suboptimal subgoals, in \Cref{fig:subgoal_seq}, we compare the subgoals produced by \(f_{\text{dc}}\) with those generated through a hand-engineered, more interpretable procedure. In the hand-engineered version, each subgoal is defined as the inventory state immediately after an intermediate item is crafted in a demonstration trajectory. The subgoal is considered achieved once the agent's inventory contains at least the same items as those target state. By sequentially completing these preset subgoals, the agent eventually crafts the ace item, as the final subgoal contains the ace item to be crafted.  Note that while such hand-crafted and interpretable subgoal decomposition is feasible for this simple TextCraft environment, in more complex settings, LLM-generated subgoal decomposition may be the only practical option.

In contrast, the subgoals proposed by \(f_{\text{dc}}\) are far less interpretable, and it is unclear how the employee agent should leverage them as effective guidance for crafting the ace item. Interestingly, despite their limited explainability and potential suboptimality, these LLM-generated subgoals still provide sufficient structure for the hierarchical agent to learn and achieve strong performance. In particular, as shown in \Cref{tab:performance}, using fully explainable, hand-engineered subgoals improves the success rate by only 2\% compared to LLM-generated ones. This suggests that, despite being less interpretable, LLM-generated subgoals still provide sufficiently strong guidance to support effective hierarchical learning.

\begin{figure}[!t]
\centering

\begin{tcolorbox}[
    colback=blue!2,
    colframe=blue!40,
    colbacktitle=blue!10,
    coltitle=black,
    title=Provided Demonstration Trajectory,
    boxrule=0.3pt,
    sharp corners,
    enhanced jigsaw,
    fonttitle=\bfseries
]
\ttfamily\raggedright
{\bfseries Goal:} craft birch trapdoor.\\[0.2em]
{\bfseries Trajectory (action, state):}\\[0.2em]

craft 6 birch slab using 3 birch planks, \char`\{\char`\}\\
get 2 birch logs, \char`\{`birch logs': 2\char`\}\\
craft 4 birch planks using 1 birch logs, \char`\{`birch planks': 4, `birch logs': 1\char`\}\\
craft 4 birch planks using 1 birch logs, \char`\{`birch planks': 8\char`\}\\
craft 2 birch trapdoor using 6 birch planks, \char`\{`birch trapdoor': 2, `birch planks': 2\char`\}
\end{tcolorbox}
\begin{tcbraster}[
  raster equal height,
  raster columns=2,
  raster width=\textwidth,
  raster column skip=3mm,
  boxrule=0.3pt,
  colback=blue!2,
  colframe=blue!40,
  colbacktitle=blue!10,
  coltitle=black,
  enhanced jigsaw,
  sharp corners,
  fonttitle=\bfseries
]

\begin{tcolorbox}[title=LLM-Generated Subgoals, width=0.45\textwidth]
\ttfamily\raggedright
\char`\{`birch logs': 1\char`\}\\
\char`\{`birch planks': 6\char`\}\\
\char`\{`birch trapdoor': 1\char`\}
\end{tcolorbox}
\begin{tcolorbox}[title=Hand-Engineered Subgoals, width=0.53\textwidth]
\ttfamily\raggedright
\char`\{`birch planks': 4, `birch logs': 1\char`\}\\
\char`\{`birch planks': 8\char`\}\\
\char`\{`birch trapdoor':\;\;2, `birch planks':\;\;2\char`\}
\end{tcolorbox}
\end{tcbraster}

\caption{Comparison of LLM-generated and hand-engineered subgoal decompositions for the demonstration trajectory shown above. Additional samples are provided in \cref{appx:llmgenVSEngSubgoal}.}
\label{fig:subgoal_seq}
\end{figure}

\begin{figure*}
\centering
\includegraphics[width=0.5\linewidth]{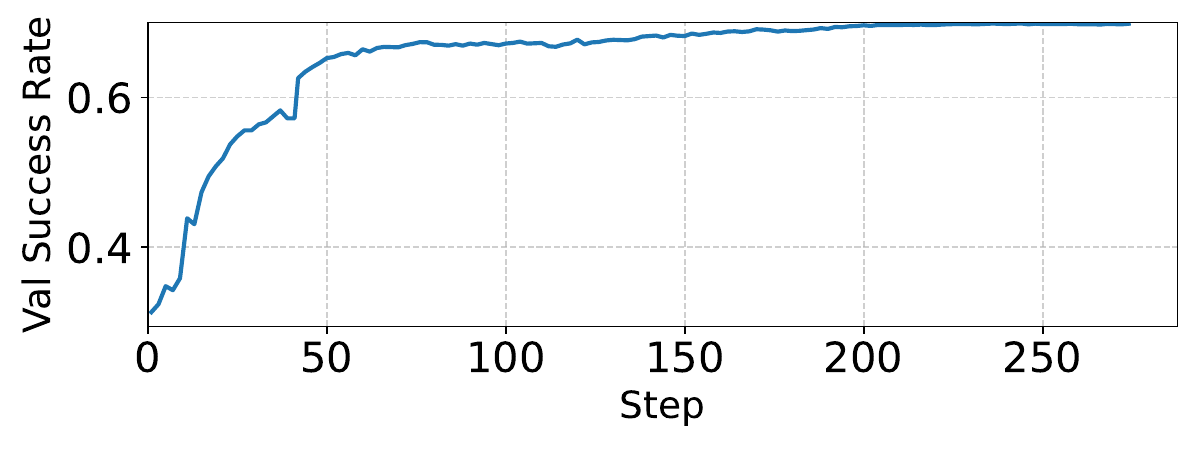}
\caption{Validation trajectory success rate for the manager agent during RL fine-tuning (Hand-engineered-subgoal). The manager progressively adapts its subgoal proposals to compensate for employee imperfections, yielding steadily improving performance.}
\label{fig:valcurves}
\end{figure*}

\subsection{Agent Imperfections and Mechanisms to Overcome Them}

Trajectory failures arise when the agent does not complete the task before exceeding the allotted step limit, and these failures may occur at either the manager or employee level.  At the manager level, failures occur when the chosen subgoals are infeasible under the available command set or are misaligned with the final objective. Moreover, even when a subgoal is valid and achievable, execution may still fail if the employee is unable to complete it due to its own imperfections.

At the employee level, failures typically stem from the agent's inability to distinguish between elementary items that must be gathered directly from the environment and intermediate items that must be crafted from them, from attempting to craft items that are infeasible given the current inventory, or from issuing invalid or unsupported crafting commands. We further observe that these errors most often arise when the employee attempts to complete a subgoal from an unfamiliar or abnormal inventory state, substantially different from those encountered during training. Such inventory states are themselves a consequence of imperfect execution of earlier subgoals, leading to situations where the agent collects irrelevant items or accumulates an excessive number of resources beyond what is needed. Notably, RL-based finetuning enables the manager to adapt to and compensate for these imperfections effectively.

\textbf{Manager Agent Accommodates Employee Imperfections.} 
To illustrate this, \Cref{tab:performance} (last row) reports a variant of our framework in which the employee is guided directly by a fixed sequence of achievable subgoals, without any manager adaptation. These subgoals are extracted from successful suboptimal trajectories in the validation set, using the same construction as the hand-engineered version (chosen in this study for its high interpretability). In principle, a perfect employee, trained on hand-engineered subgoals, could simply execute these subgoals sequentially to craft the ace item. However, without a manager to adjust the plan, this variant cannot recover when the employee fails to complete a subgoal in the sequence, resulting in a much lower success rate of 0.24. In contrast, the RL-finetuned manager adapts to such failures: when a proposed subgoal is not achieved (due to the employee's imperfect training), it receives no positive feedback and learns to propose alternative subgoals that the employee can successfully complete. Over time, the manager discovers easier, achievable subgoals that compensate for the employee's limitations, as also reflected in the steadily increasing success rates shown in \Cref{fig:valcurves}.

\begin{wrapfigure}{r}{0.45\linewidth}
    \centering
    \vspace{-15pt} 
    \includegraphics[width=\linewidth]{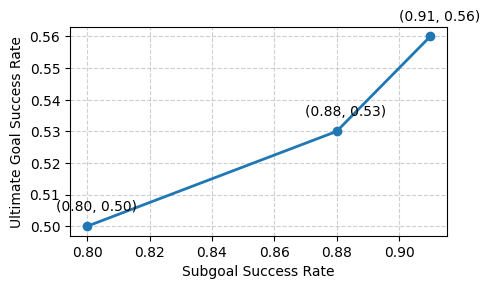}
    \caption{Subgoal vs. ultimate goal success rate in SCOPE.}
    \label{fig:SubgoalVSUltimateGoal}
    \vspace{-15pt} 
\end{wrapfigure}


\textbf{Stronger Employee Leads to Higher Success Rates.} While the manager can compensate for employee imperfections through RL finetuning, a stronger employee still substantially improves the final goal-achievement rate. \Cref{fig:SubgoalVSUltimateGoal} shows the relationship between subgoal success rate and ultimate goal success rate in SCOPE on the validation dataset, with weaker employee variants generated by evaluating partially trained checkpoints. As the figure illustrates, increases in subgoal success rate reliably translate into higher ultimate success, and this improvement accelerates when starting from a stronger employee. That is, higher subgoal success leads to disproportionately larger gains in overall goal completion. The effect is similar to compounding probabilities: the subgoal success rate applies repeatedly across steps, like a value raised to a power. When several subgoals must be achieved in sequence, even a small increase in subgoal reliability quickly compounds, resulting in much larger gains in final success.

\begin{figure*}
\centering
\includegraphics[width=0.48\linewidth]{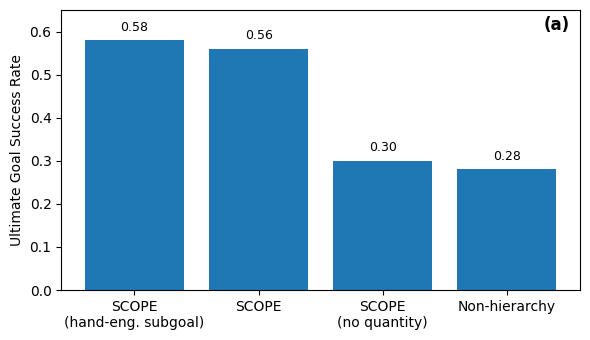}~~~
\includegraphics[width=0.48\linewidth]{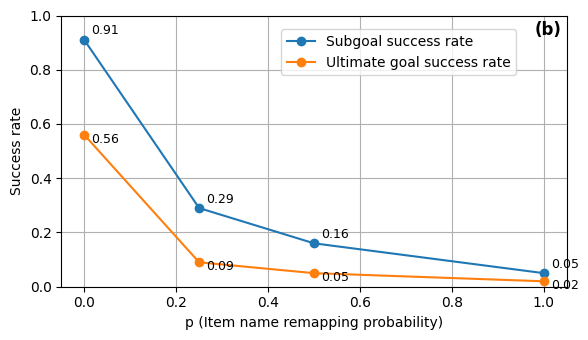}
\caption{Impact of subgoal quality on agent performance. (a) Performance vs.\ subgoal explainability (decreasing from left to right). (b) Effect of subgoals when their alignment with the environment is disrupted. Alignment is progressively broken by randomly remapping a proportion $p$ of item names in the LLM-generated subgoals, with $p \in \{0.0, 0.25, 0.5, 1.0\}$.}
\label{fig:PerformanceVSSubgoalExplanability}
\end{figure*}

\subsection{How Subgoals Help the Agent Complete the Game}
While the empirical study demonstrates that one-time LLM-generated subgoals enable SCOPE to surpass the LLM-driven hierarchical agent ADaPT, we now present additional ablations to analyze the source of this improvement more deeply.

\textbf{Impact of Subgoal Vagueness and Explainability.} As discussed in \cref{sec:impact_subopt_subgoal_performance}, LLM-generated subgoals are less explainable than hand-engineered ones, which likely contributes to the 2\% performance gap observed between SCOPE and the hand-engineered-subgoal variant. To further study how subgoal vagueness and explainability influence SCOPE’s performance, we include two additional variants, as shown in \cref{fig:PerformanceVSSubgoalExplanability}a. In SCOPE (no quantity), we remove the item-quantity information from LLM-generated subgoals, and a subgoal is considered satisfied once all listed item types have been collected at least once. Because this variant lacks quantity specifications, its subgoals are more ambiguous than those in standard SCOPE. We additionally include a limiting-case variant, Non-Hierarchical, which uses the same architecture and training setup as the employee agent (pretraining + RL fine-tuning) but is trained to pursue the final goal directly, without subgoals. In other words, the agent always conditions its decisions on the ultimate goal, rather than on any intermediate objectives. This represents the most vague scenario, as no subgoals are provided at all. As suggested by \cref{fig:PerformanceVSSubgoalExplanability}a, less interpretable and more vague subgoals make it harder for SCOPE to extract useful guidance for long-term planning, leading to a clear decline in ultimate goal success as explainability decreases.

\textbf{Effect of Decoupling Subgoals from Environment Outcomes.} To test whether subgoals contribute because they genuinely correspond to the environment outcomes, we design a mechanism that gradually breaks this connection by randomly remapping a ratio $p$ of item names in the LLM-generated subgoals. Specifically, when $p = 0$, subgoals are unchanged, which corresponds to the regular SCOPE; as $p$ increases, more item names are replaced by different ones, making the subgoals increasingly misleading; and when $p = 1.0$, all item names are remapped, meaning the subgoals no longer match the items that are actually needed in the environment. The remapping is sampled once and then fixed for both training and testing. We modify only the output of the subgoal-decomposition function, while leaving the subgoal-completion process unchanged, allowing us to degrade subgoal quality without altering the true task objective. This setup helps us understand how performance changes as the alignment between LLM-generated subgoals and environment outcomes is systematically removed, showing the extent to which correct subgoals are causally necessary for SCOPE's success.

\cref{fig:PerformanceVSSubgoalExplanability}b shows how performance changes under different remapping probabilities $p$. When $p=0$, subgoals remain unchanged, and both subgoal and ultimate success rates are those of the regular SCOPE. As we increase $p$ and disrupt the correspondence between subgoals and the environment, performance drops rapidly. At $p=0.25$, success falls to 0.29 (subgoal) and 0.09 (ultimate), and continues to decrease as remapping increases, reaching only 0.05 and 0.02 at $p=1.0$. This sharp decline confirms that SCOPE relies critically on subgoals being causally aligned with the true environmental objectives. When this alignment is lost, both subgoal success and final goal completion collapse. Notably, even with only $25\%$ of item names remapped, the ultimate success rate drops to 0.09 -- substantially lower than the non-hierarchical agent's $0.28$, which operates without subgoals at all. This suggests that when subgoals lose alignment with the environment, they do not merely become unhelpful--they can actively mislead the agent. In contrast, earlier results show that vague but aligned subgoals (e.g., using LLM-generated subgoals instead of hand-engineered ones) remain helpful and degrade performance only mildly. It is the loss of alignment--not the loss of specificity--that is particularly harmful, underscoring that subgoals must remain causally grounded to benefit the agent.

\section{Conclusion}
In this work, we introduced SCOPE, an efficient hierarchical planning framework that uses LLM-generated subgoals only once at initialization, derived from suboptimal demonstration trajectories. Instead of repeatedly querying an LLM to adaptively produce subgoals during training, SCOPE directly extracts these imperfect subgoal sequences and uses them to pretrain a student planner, followed by RL-based refinement using a world model. Although the subgoals are not optimal due to the lack of interaction between the LLM and the environment, our experiments on TextCraft show that they still provide a decent starting point for hierarchical goal decomposition. Empirically, our method outperforms the LLM-based ADaPT system \citep{PrasadKHCSBK2024}, increasing the success rate from 0.52 to 0.56. It also significantly improves efficiency: to complete the game, our agent runs in 3.0 seconds on a single NVIDIA A10 GPU, whereas ADaPT requires 164.4 seconds using a GPT-3.5 backend accessed through the OpenAI API with ideal network conditions. Overall, these results demonstrate that even suboptimal one-time LLM guidance can be highly effective when combined with RL fine-tuning, offering a practical and computationally efficient alternative to LLM-dependent hierarchical planning methods.

\newpage

\bibliographystyle{iclr2026_conference}
\bibliography{iclr2026_conference}

\appendix
\newpage    

\newpage


\section[appx]{Model architectures}
\label{appx:model_arch}
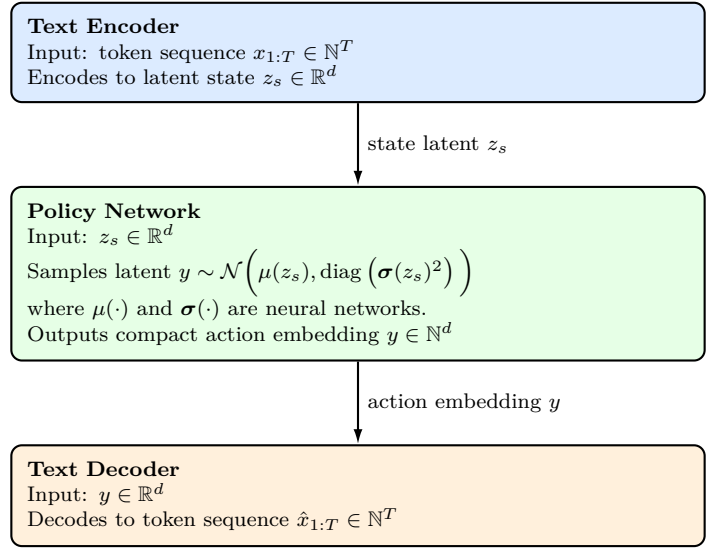
\begin{figure}[!ht]
    \centering
    
    \newcommand{\boxwidth}{0.5\linewidth}
    
    \begin{tikzpicture}[
        every node/.style={font=\footnotesize},
        box/.style={
            draw,
            rounded corners,
            thick,
            align=left,
            inner sep=6pt,
            text width=\boxwidth,
            anchor=center
        },
        arrow/.style={
            -{Latex[length=2mm,width=1.2mm]},
            thick
        },
        node distance=1.1cm
    ]
    
    \definecolor{encoderColor}{RGB}{220,235,255} 
    \definecolor{policyColor}{RGB}{230,255,230}  
    \definecolor{decoderColor}{RGB}{255,240,220} 
    
    \node[box, fill=encoderColor] (encoder) {
        \textbf{Text Encoder}\\
        Input: token sequence $x_{1:T} \in \Nb^T$\\
        Encodes to latent state $z_s \in \Rb^d$
    };
    
    \node[box, below=of encoder, fill=policyColor] (policy) {
        \textbf{Policy Network}\\
        Input: $z_s \in \Rb^d$\\
        Samples latent $y \sim \mathcal{N}\Big(\mu(z_s), \diag{\sigmav(z_s)^2}\Big)$\\
        where $\mu(\cdot)$ and $\sigmav(\cdot)$ are neural networks. \\
        Outputs compact action embedding $y \in \Nb^d$
    };
    
    \node[box, below=of policy, fill=decoderColor] (decoder) {
        \textbf{Text Decoder}\\
        Input: $y \in \Rb^d$\\
        Decodes to token sequence $\hat{x}_{1:T} \in \Nb^T$
    };
    
    \draw[arrow] (encoder.south) -- node[right]{state latent $z_s$} (policy.north);
    \draw[arrow] (policy.south) -- node[right]{action embedding $y$} (decoder.north);
    
    \end{tikzpicture}
    
    \caption{
        The Network Architecture of the employ and manager agents.
    }
    \label{fig:text-policy-architecture}
\end{figure}
Although the employee and manager agents have different input–output formats, we convert all signals into text to enable unified processing under the full text-based input–output setting. This design allows both agents to share a universal variational sequence-to-sequence architecture. Examples of output are provided in Appx.~\ref{appx:sample_output}.

As illustrated in \cref{fig:text-policy-architecture}, all networks are implemented using LSTMs \citep{HochreiterS1997}. The LSTM-based text encoder compresses the textual world-state description $x_{1:T} \!\in\! \Nb^T$ into a latent representation $z_s \!\in\! \Rb^d$. Conditioned on $z_s$, the policy network produces an action embedding $y \!\in\! \Rb^d$ via a stochastic latent variable
\[
z_a \sim \mathcal{N}\!\big(\mu(z_s), \operatorname{diag}(\sigmav(z_s)^2)\big),
\]
where $\mu(\cdot)$ and $\sigmav(\cdot)$ are neural networks, and sampling is performed using the reparameterization trick \citep{KingmaW13}. The text decoder then conditions on $y$ to generate the next action sequence $\hat{x}_{1:T} \!\in\! \Nb^T$ in natural language form.

%
\section[appx]{Hyperparameters}
\label{appx:hyperparameter}
\subsection{Autoencoder}
We train an LSTM autoencoder to obtain compact 64\text{-}dimensional latent representations of textual state-action sequences in TextCraft. Both the encoder and decoder operate over 256\text{-}dimensional token embeddings and employ a single-layer LSTM with a 512\text{-}dimensional hidden state. The model is optimized for a maximum of 2000 epochs using Adam \citep{KingmaB2015} with a learning rate of $10^{-3}$ and a batch size of 128. During training, reconstructed sequences are truncated to a maximum length of 30 tokens.

\subsection{World Model}
We employ a Transformer encoder operating over 64-dimensional latent states produced by an LSTM autoencoder. The model consists of a 2-layer Transformer with a hidden size of 256, eight attention heads, 1024-dimensional feedforward blocks, and a dropout rate of 0.1, with a \texttt{[CLS]} token used for sequence-level aggregation. Latent encodings are obtained via a single-layer LSTM autoencoder with 256-dimensional embeddings, a 512-dimensional hidden state, and a 64-dimensional bottleneck, trained using teacher forcing. Optimization is performed using Adam \citep{KingmaB2015} with a learning rate of $3\times10^{-4}$ and weight decay of $1\times10^{-5}$, together with a step-decay scheduler that halves the learning rate every 50 epochs. The world model is trained for 200~epochs with a batch size of 64, conditioning on two past states as temporal context.

\subsection{Policy Agent}
Both policy agents take as input the 64-dimensional latent states derived from the pretrained LSTM autoencoder and use a unidirectional LSTM with a 512-dimensional hidden state, 0.1 dropout. The policy is initialized for RL training from a pre-trained checkpoint and trained for a maximum of 2000 epochs with a batch size of 256 for the employee agent, and 16 for the manager agent, rolling out sequences of length 8 and using EMA smoothing ($\tau = 0.1$) together with 10\% random exploration. Training incorporates a replay buffer of capacity $10{,}000$ with reciprocal transition weighting, random replacement, and mixed sampling consisting of 90\% replayed transitions and 10\% fresh rollouts, with replay sampling enabled after $4096$ collected transitions. We retain 15\% of failed trajectories to improve robustness and perform rollouts every 10 update steps. Optimization uses Adam \citep{KingmaB2015} with a learning rate of $1\times 10^{-4}$. Action execution permits one retry per failed step, and all sequences condition on two past states as temporal context.

\newpage

\setlist[enumerate]{itemsep=1pt, topsep=2pt, parsep=0pt}

\newpage
\section[appx]{Sample outputs}
\label{appx:sample_output}
\paragraph{Trajectory 1.}
\textbf{Final goal:} \textcolor{purple}{red wool}

\textit{Available commands:}
\begin{enumerate}[leftmargin=2em]
    \item craft 1 red wool using 1 red dye, 1 white wool
    \item craft 4 magenta dye using 1 blue dye, 2 red dye, 1 white dye
    \item craft 1 loom using 2 planks, 2 string
    \item craft 1 red dye using 1 poppy
    \item craft 1 white wool using 4 string
    \item craft 2 red dye using 1 rose bush
    \item craft 8 red terracotta using 8 terracotta, 1 red dye
    \item craft 1 bow using 3 stick, 3 string
    \item craft 1 crossbow using 2 string, 3 stick, 1 iron ingot, 1 tripwire hook
    \item craft 1 red dye using 1 red tulip
    \item craft 1 green wool using 1 green dye, 1 white wool
    \item craft 1 white bed using 3 white wool, 3 planks
    \item craft 1 red dye using 1 beetroot
    \item craft 1 black wool using 1 black dye, 1 white wool
    \item craft 2 lead using 4 string, 1 slime ball
    \item craft 1 lime wool using 1 lime dye, 1 white wool
\end{enumerate}

\textbf{Policy inference:}
\begin{enumerate}[label=\textbf{Manager:}, leftmargin=2.4em, labelsep=.5em, align=left]
    \item \textcolor{blue}{Goal:} red dye (1)
    \begin{enumerate}[label=\textbf{Employee:}, leftmargin=2.8em, labelsep=.5em, align=left]
        \item \textcolor{teal}{get} 2 string
        \item \textcolor{teal}{get} 1 beetroot
        \item \textcolor{teal}{craft} 1 red dye using 1 beetroot
    \end{enumerate}

    \item \textcolor{blue}{Goal:} red dye (1), white wool (1)
    \begin{enumerate}[label=\textbf{Employee :}, leftmargin=2.8em, labelsep=.5em, align=left]
        \item \textcolor{teal}{get} 8 string
        \item \textcolor{teal}{craft} 1 white wool using 4 string
    \end{enumerate}

    \item \textcolor{blue}{Goal:} red wool (1)
    \begin{enumerate}[label=\textbf{Employee :}, leftmargin=2.8em, labelsep=.5em, align=left]
        \item \textcolor{teal}{craft} 1 red wool using 1 red dye, 1 white wool
    \end{enumerate}
\end{enumerate}

\paragraph{Trajectory 2.}
\textbf{Final goal:} \textcolor{purple}{spruce slab}

\textit{Available commands:}
\begin{enumerate}[leftmargin=2em]
    \item craft 1 spruce button using 1 spruce planks
    \item craft 4 spruce planks using 1 spruce logs
    \item craft 6 spruce slab using 3 spruce planks
    \item craft 3 spruce sign using 6 spruce planks, 1 stick
    \item craft 2 spruce trapdoor using 6 spruce planks
    \item craft 1 spruce boat using 5 spruce planks
    \item craft 1 spruce pressure plate using 2 spruce planks
    \item craft 1 spruce fence gate using 4 stick, 2 spruce planks
\end{enumerate}

\textbf{Policy inference:}
\begin{enumerate}[label=\textbf{Manager :}, leftmargin=2.4em, labelsep=.5em, align=left]
    \item \textcolor{blue}{Goal:} spruce logs (1)
    \begin{enumerate}[label=\textbf{Employee :}, leftmargin=2.8em, labelsep=.5em, align=left]
        \item \textcolor{teal}{get} 1 spruce logs
    \end{enumerate}

    \item \textcolor{blue}{Goal:} spruce planks (3)
    \begin{enumerate}[label=\textbf{Employee :}, leftmargin=2.8em, labelsep=.5em, align=left]
        \item \textcolor{teal}{craft} 4 spruce planks using 1 spruce logs
    \end{enumerate}

    \item \textcolor{blue}{Goal:} spruce slab (1)
    \begin{enumerate}[label=\textbf{Employee :}, leftmargin=2.8em, labelsep=.5em, align=left]
        \item \textcolor{teal}{craft} 6 spruce slab using 3 spruce planks
    \end{enumerate}
\end{enumerate}

\section[appx]{LLM Prompt and Output for Subgoal Proposal and Verification}
\label{appx:llm_prompt_outputs}
\subsection{Subgoal Decomposition -- \texorpdfstring{$f_\text{dc}$}{f\_dc}}

\textbf{[User]}
\begin{UserTurn}
Definition
A subgoal is defined as a high-level action or decision that, when performed correctly, significantly advances progress toward successfully completing the overall task. Subgoals may consist of one or more intermediate steps that must be achieved in order to reach the ultimate goal.

Task Description
1. Induce a consistent high-level plan or strategy-a sequence of subgoals-based on the provided set of expert trajectories.
2. Summarize the key steps required to successfully complete the task.
3. The number of subgoals should be smaller than the total number of steps in the trajectory (0 < num_subgoals < num_steps). In other words, not every step is a subgoal.
4. Use consistent methodology in determining these subgoals such that the logic is transferrable.
Output Requirements
- Produce a list of subgoals derived from the given trajectories that collectively lead to the overall goal.
- Each subgoal must be expressed in the exact format:
  the required state needed to achieve the subgoal
- The ultimate goal must appear as the final subgoal in the list.
Please complete this task systematically, so that we can later design a program to automate the process.
Input Format
The provided trajectories will have the following structure:
[goal]
[list of (action, outcome_state) tuples]
The trajectories are below:
0.
Goal: craft birch trapdoor.
craft 6 birch slab using 3 birch planks
get 2 birch logs, {'birch logs': 2}
craft 4 birch planks using 1 birch logs, {'birch planks': 4, 'birch logs': 1}
craft 4 birch planks using 1 birch logs, {'birch planks': 8}
craft 2 birch trapdoor using 6 birch planks, {'birch trapdoor': 2, 'birch planks': 2}
1.
Goal: craft black terracotta.
get 8 terracotta, {'terracotta': 8}
get 1 wither rose, {'terracotta': 8, 'wither rose': 1}
craft 1 black dye using 1 wither rose, {'terracotta': 8, 'black dye': 1}
craft 8 black terracotta using 8 terracotta, 1 black dye, {'black terracotta': 8}
2.
Goal: craft birch button.
get 1 birch logs, {'birch logs': 1}
craft 4 birch planks using 1 birch logs, {'birch planks': 4}
craft 1 birch button using 1 birch planks, {'birch planks': 3, 'birch button': 1}
3.
Goal: craft pink banner.
(get 6 pink tulip ... craft 1 pink banner using 6 pink wool, 1 stick)
4.
Goal: craft crossbow.
(get 2 string ... craft 1 crossbow using 2 string, 3 stick, 1 iron ingot, 1 tripwire hook)
\end{UserTurn}

\textbf{[GPT-5]}
\begin{ModelTurn}
Here’s a consistent, recipe-aware decomposition for each trajectory...

# 0) Goal: craft birch trapdoor
- {'birch logs': 1}
- {'birch planks': 6}
- {'birch trapdoor': 1}

# 1) Goal: craft black terracotta
- {'terracotta': 8}
- {'black dye': 1}
- {'black terracotta': 8}

# 2) Goal: craft birch button
- {'birch logs': 1}
- {'birch planks': 1}
- {'birch button': 1}

# 3) Goal: craft pink banner
- {'white wool': 6}
- {'pink dye': 6}
- {'pink wool': 6}
- {'stick': 1}
- {'pink banner': 1}

# 4) Goal: craft crossbow
- {'string': 2}
- {'stick': 4}
- {'iron ingot': 2}
- {'warped planks': 1}
- {'tripwire hook': 1}
- {'crossbow': 1}
\end{ModelTurn}

\textbf{[User]}
\begin{UserTurn}
Good, now create a Python function to programmatically perform this inference
\end{UserTurn}

\textbf{[GPT-5]}
\begin{ModelTurn}
import re
from collections import defaultdict, deque

CRAFT_RE = re.compile(r"^craft\s+(\d+)\s+([a-z ]+?)\s+using\s+(.+)$")
GET_RE = re.compile(r"^get\s+(\d+)\s+([a-z ]+?)$")

def _parse_ingredients(txt):
    """
    '8 terracotta, 1 black dye' -> list of (qty:int, item:str)
    """
    parts = [p.strip() for p in txt.split(",")]
    out = []
    for p in parts:
        m = re.match(r"^(\d+)\s+([a-z ]+?)$", p.strip())
        if not m:
            # be tolerant of stray punctuation (e.g., trailing periods)
            p = p.strip().rstrip(".,;:")
            m = re.match(r"^(\d+)\s+([a-z ]+?)$", p)
        if m:
            out.append((int(m.group(1)), m.group(2).strip()))
    return out

def _parse_action(line):
    """
    Returns a dict with keys:
      kind: 'craft' or 'get' or None
      out_qty, out_item, ing: for craft
      qty, item: for get
    """
    line = line.strip()
    # cut off trailing inventory dict if present: ", {...}"
    if ", {" in line:
        line = line[:line.index(", {")].strip()

    m = CRAFT_RE.match(line)
    if m:
        out_qty = int(m.group(1))
        out_item = m.group(2).strip()
        ing_txt = m.group(3).strip()
        # tolerate a trailing inventory after a comma already removed; also strip trailing periods
        ing_txt = ing_txt.rstrip(".")
        ing = _parse_ingredients(ing_txt)
        return {"kind": "craft", "out_qty": out_qty, "out_item": out_item, "ing": ing}

    m = GET_RE.match(line)
    if m:
        return {"kind": "get", "qty": int(m.group(1)), "item": m.group(2).strip()}

    return {"kind": None}

def infer_subgoals_from_trajectory(goal_item, action_lines):
    """
    Parameters
    ----------
    goal_item : str
        e.g., 'birch trapdoor', 'black terracotta'
    action_lines : list[str]
        Lines like:
          'get 8 terracotta, {...}'
          'craft 8 black terracotta using 8 terracotta, 1 black dye, {...}'

    Returns
    -------
    subgoals : list[dict]
        Each element is a required inventory state (e.g., {'terracotta': 8})
        Ordered from base → intermediates → final goal.
    """
    # 1) Parse actions into recipes and note which items can be crafted.
    recipes = {}  # product_item -> (out_qty, [(qty, item), ...])
    seen_get = set()
    all_actions = []
    for line in action_lines:
        a = _parse_action(line)
        all_actions.append(a)
        if a["kind"] == "craft":
            recipes[a["out_item"]] = (a["out_qty"], a["ing"])
        elif a["kind"] == "get":
            seen_get.add(a["item"])

    # 2) Find the final craft that produces the goal_item (if present).
    # If absent, we still assume we want at least 1 goal_item.
    if goal_item not in recipes:
        # We can't expand dependencies; just return the final subgoal.
        return [{goal_item: 1}]

    # 3) Recursive expansion to compute ingredient requirements for one goal item.
    # We will compute a dependency graph and a "required counts" per node.
    # For items with recipes, we compute how much of each ingredient is needed
    # to yield the requested quantity (ceil div if needed).
    import math

    # store required quantities for intermediates (not bases)
    required = defaultdict(int)
    # frontier items to expand: (item, qty_needed)
    frontier = deque([(goal_item, 1)])

    # Graph for topo order: edges (ingredient -> product)
    edges = defaultdict(set)
    nodes = set()

    # Track per-item total requirement (for all nodes, including base)
    total_required = defaultdict(int)
    total_required[goal_item] = 1

    while frontier:
        item, need_qty = frontier.popleft()
        nodes.add(item)

        if item in recipes:
            out_qty, ing = recipes[item]
            # crafts needed to produce at least need_qty
            crafts_needed = math.ceil(need_qty / out_qty)
            # accumulate requirements for ingredients
            for q, ing_item in ing:
                req = q * crafts_needed
                total_required[ing_item] += req
                edges[ing_item].add(item)
                nodes.add(ing_item)

                # If ingredient itself is craftable, expand further; else it's base.
                if ing_item in recipes:
                    # push what *this node* needs (not cumulative to avoid overcounting)
                    frontier.append((ing_item, req))

    # 4) Identify base vs intermediates (base = no recipe)
    base_items = [n for n in nodes if n not in recipes and n != goal_item]
    inter_items = [n for n in nodes if n in recipes and n != goal_item]

    # 5) Topological order (Kahn) so that base → intermediates → goal.
    indeg = {n: 0 for n in nodes}
    for u in edges:
        for v in edges[u]:
            indeg[v] += 1

    q = deque([n for n in nodes if indeg[n] == 0])
    topo = []
    while q:
        u = q.popleft()
        topo.append(u)
        for v in edges.get(u, []):
            indeg[v] -= 1
            if indeg[v] == 0:
                q.append(v)

    # 6) Build subgoals:
    #    - For each base item appearing before any of its consumers, include {item: required_count}
    #    - For each intermediate item, include {item: required_count}
    #    - End with {goal_item: 1}
    subgoals = []
    for n in topo:
        if n == goal_item:
            continue
        if n in total_required and total_required[n] > 0:
            subgoals.append({n: total_required[n]})

    # Final goal state: have at least 1 of the goal item.
    subgoals.append({goal_item: 1})
    return subgoals
\end{ModelTurn}

\subsection{Subgoal Completion -- \texorpdfstring{$f_\text{sg}$}{f\_\text{sg}}}

\textbf{[User]}
\begin{UserTurn}
Definition
A subgoal is defined as a high-level action or decision that, when performed correctly, significantly advances progress toward successfully completing the overall task. Subgoals may consist of one or more intermediate steps that must be achieved in order to reach the ultimate goal.

Task Description
1. Induce a consistent high-level plan or strategy--a sequence of subgoals--based on the provided set of expert trajectories.
2. Summarize the key steps required to successfully complete the task.
3. The number of subgoals should be smaller than the total number of steps in the trajectory (0 < num_subgoals < num_steps). In other words, not every step is a subgoal.
4. Use consistent methodology in determining these subgoals such that the logic is transferrable.
Output Requirements
- Produce a list of subgoals derived from the given trajectories that collectively lead to the overall goal.
- Each subgoal must be expressed in the exact format:
  the required state needed to achieve the subgoal
- The ultimate goal must appear as the final subgoal in the list.
- Add a reward function such that, given a current state, returns whether a proposed subgoal has been achieved or not (binary label)
Please complete this task systematically, so that we can later design a program to automate the process.
Input Format
The provided trajectories will have the following structure:
[goal]
[list of (action, outcome_state) tuples]
The trajectories are below:
0.
Goal: craft birch trapdoor.
craft 6 birch slab using 3 birch planks
get 2 birch logs, {'birch logs': 2}
craft 4 birch planks using 1 birch logs, {'birch planks': 4, 'birch logs': 1}
craft 4 birch planks using 1 birch logs, {'birch planks': 8}
craft 2 birch trapdoor using 6 birch planks, {'birch trapdoor': 2, 'birch planks': 2}
1.
Goal: craft black terracotta.
get 8 terracotta, {'terracotta': 8}
get 1 wither rose, {'terracotta': 8, 'wither rose': 1}
craft 1 black dye using 1 wither rose, {'terracotta': 8, 'black dye': 1}
craft 8 black terracotta using 8 terracotta, 1 black dye, {'black terracotta': 8}
2.
Goal: craft birch button.
get 1 birch logs, {'birch logs': 1}
craft 4 birch planks using 1 birch logs, {'birch planks': 4}
craft 1 birch button using 1 birch planks, {'birch planks': 3, 'birch button': 1}
3.
Goal: craft pink banner.
get 6 pink tulip, {'pink tulip': 6}
craft 1 pink dye using 1 pink tulip, {'pink tulip': 5, 'pink dye': 1}
get 3 bone meal, {'pink tulip': 5, 'pink dye': 1, 'bone meal': 3}
craft 4 birch planks using 1 birch logs, {'pink tulip': 5, 'pink dye': 1, 'bone meal': 3, 'birch planks': 4}
craft 1 pink dye using 1 pink tulip, {'pink tulip': 4, 'pink dye': 2, 'bone meal': 3, 'birch planks': 4}
craft 1 pink dye using 1 pink tulip, {'pink tulip': 3, 'pink dye': 3, 'bone meal': 3, 'birch planks': 4}
craft 1 pink dye using 1 pink tulip, {'pink tulip': 2, 'pink dye': 4, 'bone meal': 3, 'birch planks': 4}
craft 1 pink dye using 1 pink tulip, {'pink tulip': 1, 'pink dye': 5, 'bone meal': 3, 'birch planks': 4}
craft 1 pink dye using 1 pink tulip, {'pink tulip': 0, 'pink dye': 6, 'bone meal': 3, 'birch planks': 4}
get 24 string, {'pink dye': 6, 'bone meal': 3, 'birch planks': 4, 'string': 24}
craft 1 white wool using 4 string, {'pink dye': 6, 'bone meal': 3, 'birch planks': 4, 'string': 20, 'white wool': 1}
craft 1 white wool using 4 string, {'pink dye': 6, 'bone meal': 3, 'birch planks': 4, 'string': 16, 'white wool': 2}
craft 1 white wool using 4 string, {'pink dye': 6, 'bone meal': 3, 'birch planks': 4, 'string': 12, 'white wool': 3}
craft 1 white wool using 4 string, {'pink dye': 6, 'bone meal': 3, 'birch planks': 4, 'string': 8, 'white wool': 4}
craft 1 white wool using 4 string, {'pink dye': 6, 'bone meal': 3, 'birch planks': 4, 'string': 4, 'white wool': 5}
craft 1 white wool using 4 string, {'pink dye': 6, 'bone meal': 3, 'birch planks': 4, 'string': 0, 'white wool': 6}
craft 1 pink wool using 1 pink dye, 1 white wool, {'pink dye': 5, 'bone meal': 3, 'birch planks': 4, 'white wool': 5, 'pink wool': 1}
craft 1 pink wool using 1 pink dye, 1 white wool, {'pink dye': 4, 'bone meal': 3, 'birch planks': 4, 'white wool': 4, 'pink wool': 2}
craft 1 pink wool using 1 pink dye, 1 white wool, {'pink dye': 3, 'bone meal': 3, 'birch planks': 4, 'white wool': 3, 'pink wool': 3}
craft 1 pink wool using 1 pink dye, 1 white wool, {'pink dye': 2, 'bone meal': 3, 'birch planks': 4, 'white wool': 2, 'pink wool': 4}
craft 1 pink wool using 1 pink dye, 1 white wool, {'pink dye': 1, 'bone meal': 3, 'birch planks': 4, 'white wool': 1, 'pink wool': 5}
craft 1 pink wool using 1 pink dye, 1 white wool, {'pink dye': 0, 'bone meal': 3, 'birch planks': 4, 'white wool': 0, 'pink wool': 6}
get 2 bamboo, {'bone meal': 3, 'birch planks': 4, 'pink wool': 6, 'bamboo': 2}
craft 1 stick using 2 bamboo, {'bone meal': 3, 'birch planks': 4, 'pink wool': 6, 'bamboo': 0, 'stick': 1}
craft 1 pink banner using 6 pink wool, 1 stick, {'bone meal': 3, 'birch planks': 4, 'pink banner': 1}
4.
Goal: craft crossbow.
get 2 string, {'string': 2}
get 6 bamboo, {'string': 2, 'bamboo': 6}
craft 1 stick using 2 bamboo, {'string': 2, 'bamboo': 4, 'stick': 1}
craft 4 spruce planks using 1 spruce logs, {'string': 2, 'bamboo': 4, 'stick': 1, 'spruce planks': 4}
craft 1 stick using 2 bamboo, {'string': 2, 'bamboo': 2, 'stick': 2, 'spruce planks': 4}
get 4 spruce planks, {'string': 2, 'bamboo': 2, 'stick': 2, 'spruce planks': 8}
craft 1 stick using 2 bamboo, {'string': 2, 'bamboo': 0, 'stick': 3, 'spruce planks': 8}
get 1 iron ingot, {'string': 2, 'bamboo': 0, 'stick': 3, 'spruce planks': 8, 'iron ingot': 1}
get 2 bamboo, {'string': 2, 'bamboo': 2, 'stick': 3, 'spruce planks': 8, 'iron ingot': 1}
craft 1 stick using 2 bamboo, {'string': 2, 'bamboo': 0, 'stick': 4, 'spruce planks': 8, 'iron ingot': 1}
get 1 iron ingot, {'string': 2, 'bamboo': 0, 'stick': 4, 'spruce planks': 8, 'iron ingot': 2}
get 1 warped stems, {'string': 2, 'bamboo': 0, 'stick': 4, 'spruce planks': 8, 'iron ingot': 2, 'warped stems': 1}
craft 4 warped planks using 1 warped stems, {'string': 2, 'bamboo': 0, 'stick': 4, 'spruce planks': 8, 'iron ingot': 2, 'warped planks': 4}
craft 2 tripwire hook using 1 stick, 1 iron ingot, 1 warped planks, {'string': 2, 'bamboo': 0, 'stick': 3, 'spruce planks': 8, 'iron ingot': 1, 'warped planks': 3, 'tripwire hook': 2}
craft 1 crossbow using 2 string, 3 stick, 1 iron ingot, 1 tripwire hook, {'spruce planks': 8, 'warped planks': 3, 'tripwire hook': 1, 'crossbow': 1}
\end{UserTurn}

\textbf{[GPT-5]}
\begin{ModelTurn}
Got it -- here's a clean, transferable method plus the concrete subgoals for each trajectory and a tiny reward checker you can drop into your pipeline.

# Method (used consistently below)
1) Read the final recipe and list its direct ingredients.  
2) For any ingredient that is itself crafted, compress its internal chain into a single milestone state (e.g., "have 6 pink wool"), rather than exposing every micro-step.  
3) Keep only necessary inventory thresholds that enable the next craft.  
4) Ensure 0 < #subgoals < #steps and end with the ultimate goal state (>=1 of target item).

---

## 0) Goal: craft birch trapdoor
**Key steps (summary):** convert birch logs -> planks; once you have >=6 birch planks, craft trapdoor.

**Subgoals (states only, final one is the goal):**
1. {'birch planks': 6}  
2. {'birch trapdoor': 1}

---

## 1) Goal: craft black terracotta
**Key steps (summary):** obtain terracotta (8); obtain a black dye (e.g., from wither rose); craft black terracotta.

**Subgoals:**
1. {'terracotta': 8}  
2. {'black dye': 1}  
3. {'black terracotta': 1}

---

## 2) Goal: craft birch button
**Key steps (summary):** make birch planks; craft button.

**Subgoals:**
1. {'birch planks': 1}  
2. {'birch button': 1}

---

## 3) Goal: craft pink banner
**Key steps (summary):** produce 6 pink wool (via white wool + pink dye), get a stick, craft banner.

**Subgoals:**
1. {'pink wool': 6}  
2. {'stick': 1}  
3. {'pink banner': 1}

*(Note: This compresses the internal chains "string -> white wool" and "tulip -> pink dye -> pink wool" into the single milestone `pink wool: 6`.)*

---

## 4) Goal: craft crossbow
**Key steps (summary):** craft a tripwire hook; ensure remaining ingredients (2 string, 3 sticks, 1 iron ingot) are on hand; craft crossbow.

**Subgoals:**
1. {'tripwire hook': 1}  
2. {'string': 2, 'stick': 3, 'iron ingot': 1}  
3. {'crossbow': 1}

*(Note: The hook's own chain (stick + iron ingot + planks) is compressed into the milestone ``have 1 tripwire hook''.)*

---

# Binary Reward Function (subgoal achieved?)
from typing import Dict

def subgoal_achieved(current_state: Dict[str, int], subgoal_state: Dict[str, int]) -> int:
    """
    Returns 1 if current_state satisfies (>=) every item requirement in subgoal_state; else 0.
    Both dicts map item_name -> nonnegative int quantity.
    """
    for item, req_qty in subgoal_state.items():
        if current_state.get(item, 0) < req_qty:
            return 0
    return 1

### Example usage
# Example for trajectory 3 (pink banner)
current = {'pink wool': 6, 'stick': 0}
sg1 = {'pink wool': 6}
sg2 = {'stick': 1}
sg3 = {'pink banner': 1}

assert subgoal_achieved(current, sg1) == 1
assert subgoal_achieved(current, sg2) == 0
assert subgoal_achieved({'pink banner': 1}, sg3) == 1

If you want, I can package these into a JSON schema (trajectories -> subgoals array) or wire the checker into your rollout collector.
\end{ModelTurn}

\newpage

\subsection{Comparison of LLM-generated and Hand-engineered Subgoal Decompositions}
\label{appx:llmgenVSEngSubgoal}

\subsubsection{Trajectory 1: Crafting Black Terracotta}

\begin{tcolorbox}[
    colback=blue!2,
    colframe=blue!40,
    colbacktitle=blue!10,
    coltitle=black,
    title=Provided Demonstration Trajectory,
    boxrule=0.3pt,
    sharp corners,
    enhanced jigsaw,
    fonttitle=\bfseries
]
\ttfamily\raggedright
{\bfseries Goal:} \;\;craft black terracotta.\\[0.2em]
{\bfseries Trajectory (action, state):}\\[0.2em]
get 8 terracotta, \char`\{`terracotta':\;\;8\char`\}\\
get 1 wither rose, \char`\{`terracotta':\;\;8, `wither rose':\;\;1\char`\}\\
craft 1 black dye using 1 wither rose, \char`\{`terracotta':\;\;8, `black dye':\;\;1\char`\}\\
craft 8 black terracotta using 8 terracotta, 1 black dye, \char`\{`black terracotta':\;\;8\char`\}
\end{tcolorbox}

\vspace{0.5em}

\begin{tcbraster}[
  raster equal height,
  raster columns=2,
  raster width=\textwidth,
  raster column skip=3mm,
  boxrule=0.3pt,
  colback=blue!2,
  colframe=blue!40,
  colbacktitle=blue!10,
  coltitle=black,
  enhanced jigsaw,
  sharp corners,
  fonttitle=\bfseries
]

\begin{tcolorbox}[title=LLM-Generated Subgoals]
\ttfamily\raggedright
\char`\{`terracotta':\;\;8\char`\}\\
\char`\{`black dye':\;\;1\char`\}\\
\char`\{`black terracotta':\;\;8\char`\}
\end{tcolorbox}
\begin{tcolorbox}[title=Hand-Engineered Subgoals]
\ttfamily\raggedright
\char`\{`terracotta':\;\;8, `black dye':\;\;1\char`\}\\
\char`\{`black terracotta':\;\;8\char`\}
\end{tcolorbox}

\end{tcbraster}

\subsubsection{Trajectory 2: Crafting Birch Button}

\begin{tcolorbox}[
    colback=blue!2,
    colframe=blue!40,
    colbacktitle=blue!10,
    coltitle=black,
    title=Provided Demonstration Trajectory,
    boxrule=0.3pt,
    sharp corners,
    enhanced jigsaw,
    fonttitle=\bfseries
]
\ttfamily\raggedright
{\bfseries Goal:} \;\;craft birch button.\\
{\bfseries Trajectory (action, state):}\\[0.2em]
get 1 birch logs, \char`\{`birch logs':\;\;1\char`\}\\
craft 4 birch planks using 1 birch logs, \char`\{`birch planks':\;\;4\char`\}\\
craft 1 birch button using 1 birch planks, \char`\{`birch planks':\;\;3, `birch button':\;\;1\char`\}
\end{tcolorbox}

\vspace{0.5em}

\begin{tcbraster}[
  raster equal height,
  raster columns=2,
  raster width=\textwidth,
  raster column skip=3mm,
  boxrule=0.3pt,
  colback=blue!2,
  colframe=blue!40,
  colbacktitle=blue!10,
  coltitle=black,
  enhanced jigsaw,
  sharp corners,
  fonttitle=\bfseries
]

\begin{tcolorbox}[title=LLM-Generated Subgoals]
\ttfamily\raggedright
\char`\{`birch logs':\;\;1\char`\}\\
\char`\{`birch planks':\;\;1\char`\}\\
\char`\{`birch button':\;\;1\char`\}
\end{tcolorbox}
\begin{tcolorbox}[title=Hand-Engineered Subgoals]
\ttfamily\raggedright
\char`\{`birch planks':\;\;4\char`\}\\
\char`\{`birch button':\;\;1, `birch planks':\;\;3\char`\}
\end{tcolorbox}

\end{tcbraster}

\subsubsection{Trajectory 3: Crafting Crossbow}

\begin{tcolorbox}[
    colback=blue!2,
    colframe=blue!40,
    colbacktitle=blue!10,
    coltitle=black,
    title=Provided Demonstration Trajectory,
    boxrule=0.3pt,
    sharp corners,
    enhanced jigsaw,
    fonttitle=\bfseries
]
\ttfamily\raggedright
{\bfseries Goal:} \;\;craft crossbow.\\[0.2em]
{\bfseries Trajectory (action, state):}\\[0.2em]
get 2 string, \char`\{`string':\;\;2\char`\}\\
get 6 bamboo, \char`\{`string':\;\;2, `bamboo':\;\;6\char`\}\\
craft 1 stick using 2 bamboo, \char`\{`string':\;\;2, `bamboo':\;\;4, `stick':\;\;1\char`\}\\
craft 4 spruce planks using 1 spruce logs, \char`\{`string':\;\;2, `bamboo':\;\;4, `stick':\;\;1\char`\}\\
craft 1 stick using 2 bamboo, \char`\{`string':\;\;2, `bamboo':\;\;2, `stick':\;\;2\char`\}\\
get 4 spruce planks, \char`\{`string':\;\;2, `bamboo':\;\;2, `stick':\;\;2\char`\}\\
craft 1 stick using 2 bamboo, \char`\{`string':\;\;2, `stick':\;\;3\char`\}\\
get 1 iron ingot, \char`\{`string':\;\;2, `stick':\;\;3, `iron ingot':\;\;1\char`\}\\
get 2 bamboo, \char`\{`string':\;\;2, `bamboo':\;\;2, `stick':\;\;3, `iron ingot':\;\;1\char`\}\\
craft 1 stick using 2 bamboo, \char`\{`string':\;\;2, `stick':\;\;4, `iron ingot':\;\;1\char`\}\\
get 1 iron ingot, \char`\{`string':\;\;2, `stick':\;\;4, `iron ingot':\;\;2\char`\}\\
get 1 warped stems, \char`\{`string':\;\;2, `stick':\;\;4, `iron ingot':\;\;2, `warped stems':\;\;1\char`\}\\
craft 4 warped planks using 1 warped stems, \char`\{`string':\;\;2, `stick':\;\;4, `iron ingot':\;\;2, `warped planks':\;\;4\char`\}\\
craft 2 tripwire hook using 1 stick, 1 iron ingot, 1 warped planks, \char`\{`string':\;\;2, `stick':\;\;3, `iron ingot':\;\;1, `warped planks':\;\;3, `tripwire hook':\;\;2\char`\}\\
craft 1 crossbow using 2 string, 3 stick, 1 iron ingot, 1 tripwire hook, \char`\{`warped planks':\;\;3, `tripwire hook':\;\;1, `crossbow':\;\;1\char`\}
\end{tcolorbox}

\vspace{0.5em}

\begin{tcbraster}[
  raster equal height,
  raster columns=2,
  raster width=\textwidth,
  raster column skip=3mm,
  boxrule=0.3pt,
  colback=blue!2,
  colframe=blue!40,
  colbacktitle=blue!10,
  coltitle=black,
  enhanced jigsaw,
  sharp corners,
  fonttitle=\bfseries
]

\begin{tcolorbox}[title=LLM-Generated Subgoals]
\ttfamily\raggedright
\char`\{`string':\;\;2\char`\}\\
\char`\{`stick':\;\;4\char`\}\\
\char`\{`iron ingot':\;\;2\char`\}\\
\char`\{`warped planks':\;\;1\char`\}\\
\char`\{`tripwire hook':\;\;1\char`\}\\
\char`\{`crossbow':\;\;1\char`\}
\end{tcolorbox}
\begin{tcolorbox}[title=Hand-Engineered Subgoals]
\ttfamily\raggedright
\char`\{`string':\;\;2, `bamboo':\;\;4, `stick':\;\;1\char`\}\\
\char`\{`string':\;\;2, `bamboo':\;\;2, `stick':\;\;2\char`\}\\
\char`\{`string':\;\;2,`stick':\;\;3\char`\}\\
\char`\{`string':\;\;2, `iron ingot':\;\;1, `stick':\;\;4\char`\}\\
\char`{`string':\;\;2, `stick':\;\;4, `iron ingot':\;\;2, `warped planks':\;\;4\char`\}\\
\char`\{`string':\;\;2, `stick':\;\;3, `iron ingot':\;\;1, `warped planks':\;\;3, `tripwire hook':\;\;2\char`\}\\
 \char`\{`warped planks':\;\;3, `tripwire hook':\;\;1, `crossbow':\;\;1\char`\}
\end{tcolorbox}

\end{tcbraster}



\end{document}